\documentclass{article}

\usepackage{microtype}
\usepackage{graphicx}
\usepackage{subfigure}
\usepackage{booktabs} 
\usepackage{amssymb}
\usepackage{amsmath}
\usepackage{xcolor}
\usepackage{amsthm}
\usepackage{mathtools}
\usepackage{multirow}
\usepackage{array}

\newcommand{\ignore}[1]{}

\newcommand{\sumiN}{\sum_{j=1}^{n}}

\newcommand{\etazone}{\eta_{{FR}}(z_j = 1)}
\newcommand{\etazzero}{\eta_{{FR}}(z_j = 0)}

\newcommand{\etaczone}{\eta_{{FC}}(z_j = 1)}
\newcommand{\etaczzero}{\eta_{{FC}}(z_j = 0)}

\newcommand{\loglossobj}{\frac{1}{m} \sum_{i=1}^m \log \Big (1+\exp(-y_iA_i x) \Big )}
\newcommand{\logloss}{{\mathcal L}(x)}

\newcommand{\mipreg}{\eqref{eq: persp_reg_mip}}
\newcommand{\mipcard}{\eqref{eq: persp_card_mip}}

\newcommand{\perspreg}{\eqref{eq: persp_reg}}
\newcommand{\perspcard}{\eqref{eq: persp_card}}

\newcommand{\fdr}{\eqref{eq: reg_fd}}
\newcommand{\fdc}{\eqref{eq: card_fd}}

\newcommand{\proponepenalty}{\mu-\gamma\delta_j}
\theoremstyle{plain}
\newtheorem{prop}{Proposition}

\usepackage{tabularx}

\usepackage{hyperref}

\usepackage[accepted]{icml2022}
\usepackage[colorinlistoftodos,prependcaption,textsize=small]{todonotes}

\icmltitlerunning{Safe Screening for Logistic Regression with $\ell_0$--$\ell_2$ Regularization}

\begin{document}

\twocolumn[
\icmltitle{Safe Screening for Logistic Regression with $\ell_0$--$\ell_2$ Regularization}



\icmlsetsymbol{equal}{*}

\begin{icmlauthorlist}
\icmlauthor{Anna Deza}{ucb}
\icmlauthor{Alper Atamt\"urk}{ucb}
\end{icmlauthorlist}

\icmlaffiliation{ucb}{Department of Indutrial Engineering and Operations Research, University of California, Berkeley, CA, USA}

\icmlcorrespondingauthor{Alper Atamt\"urk}{atamturk@berkeley.edu}

\icmlkeywords{Machine Learning, ICML}

\vskip 0.3in
]



\printAffiliationsAndNotice{}  

\begin{abstract}
In logistic regression, it is often desirable to utilize regularization to promote sparse solutions, particularly for problems with a large number of features compared to available labels. 
In this paper,
we present screening rules that safely remove features from logistic regression with $\ell_0-\ell_2$ regularization before solving the problem. The proposed safe screening rules are based on lower bounds from the Fenchel dual of strong conic relaxations of the logistic regression problem. Numerical experiments with real and synthetic data suggest that a high percentage of the features can be effectively and safely removed apriori, leading to substantial speed-up in the computations. 
\end{abstract}

\section{Introduction}
Logistic regression is a classification model used to predict the probability of a binary outcome from a set of features.
Its use is prevalent in a large variety of domains, from diagnostics in healthcare \citep{gramfort2013time, shevade2003simple, cawley2006gene} to sentiment analysis in natural language processing \cite{wang2017active, yen2011logistic} and consumer choice modeling in economics \cite{kuswanto2015logistic}. 

Given a data matrix 
$A \in \mathbb{R}^{m \times n}$ of $m$ observations, each with $n$ features and binary labels $y \in \{-1, 1\}^m$, 
the logistics regression model seeks regression coefficients $x \in \mathbb{R}^n$ that minimize the convex loss function
$$\logloss = \loglossobj \cdot$$ 
We use $A_i$ to denote the $i\text{-th}$ row of matrix $A$ and $A^j$ to denote the $j$-th column of $A$. 
When the number of available features is large compared to the number of the observations (labels),
i.e., $ m << n$, logistic regression models are prone to overfitting. Such cases require pruning the features to mitigate the risk of overfitting. Regularization is a natural approach for this purpose.
Convex $\ell_2$-regularization (ridge) \cite{hoerl1970ridge} imposes bias by shrinking the regression coefficients $x_i$, $ i \in [n],$ toward zero. The $\ell_1$-regularization (lasso) \cite{tibshirani1996regression} and $\ell_1$--$\ell_2$-regularization (elastic net) \cite{zou2005regularization} perform shrinkage of the coefficients and selection of the features simultaneously. Recently, there has been a growing interested in
utilizing the exact $\ell_0$-regularization \cite{miller2002subset,bertsimas2016best} for selecting features in linear regression. Although $\ell_0$-regularization introduces non-convexity to regression models, significant progress has been done to develop strong models and specialized algorithms to solve medium to large scale instances recently \citep[e.g.][]{bertsimas2017sparse,AG:rank-one,hazimeh2020fast,HGA:2x2}.

We consider logistic regression with $\ell_0$--$\ell_2$ regularization:
\begin{align*}
    &\min_{x\in\mathbb{R}^n} \logloss + \frac{1}{\gamma} \|x\|_2^2 + \mu \|x\|_0 \text {, and } \quad & \text{(REG)} \nonumber \label{REG}\\
    &\min_{x\in\mathbb{R}^n} \logloss + \frac{1}{\gamma} \|x\|_2^2  \ \text{ s.t. } \|x\|_0 \leq k. \quad & \text{(CARD)} \nonumber 
\end{align*}
Whereas the $\ell_2$-regularization penalty term above encourages shrinking the coefficients, which helps counter effects of noise present in the data matrix $A$, the $\ell_0$-regularization penalty term in (REG) encourages sparsity, selecting a small number of key features to be used for prediction, which is modeled as an explicit cardinality constraint in (CARD). Due to the $\ell_0$-regularization terms, (REG) and (CARD) are non-convex optimization problems.

Screening rules refer to preprocessing procedures that discard certain features, leading to a reduction in the dimension of the problem, which, in turn, improves the solution times of the employed algorithms. For $\ell_1$-regularized linear regression, \citet{ghaoui2010safe} introduce safe screening rules that guarantee to remove only features that are not selected in the solution. Strong rules \cite{tibshirani2011regression}, on the other hand, are heuristics with no guarantee but able to prune a large number of features fast. A large body of work exists on screening rules for $\ell_1$-regularized regression \cite{wang2013lasso, liu2014safe, fercoq2015mind, ndiaye2017gap, dantas2021expanding}, including some for logistic regression \cite{wang2014safe}. However, little attention has been given to the $\ell_0$-regularized regression problem, where dimension reduction by screening rules can have substantially larger impact due to the higher computational burden for solving the non-convex regression problems.
Bounds from strong conic relaxations of $\ell_0$-regularized problems \cite{atamturk2018sparse,AG:rank-one} substantially
reduce the computational burden with effective pruning strategies.
Recently, \citet{atamturk2020safe} propose safe screening rules for the $\ell_0$-regularized linear regression problem from perspective 
relaxations. To the best of our knowledge, no screening rule exists in the literature for the logistic regression problems (REG) and (CARD) with $\ell_0$--$\ell_2$ regularization, studied in this paper.

\textbf{Outline} In Section~\ref{sec:conic}, we give strong conic mixed 0-1 formulations for logistic regression problems (REG) and (CARD) with $\ell_0$--$\ell_2$ regularization.
In Section~\ref{sec:screen}, we derive the safe screening rules for them based on bounds from Fenchel duals of their conic relaxations and in Section~\ref{sec:comp}, we summarize the computational experiments  performed for testing the effectiveness of the proposed screening rules for $\ell_0$--$\ell_2$ logistic regression problems with synthetic as well as real data. Finally, we conclude with a few final remarks in Section~\ref{sec:conc}.




\section{Conic Reformulations}
\label{sec:conic}

In this section, we present strong conic formulations for (REG) and (CARD).
First, we state convex logistic regression loss minimization as 
a conic optimization problem.
Writing the epigraph of the softplus function
$\log (1+ \exp(x)) \le s$ as an upper bound on the 
sum of two exponential functions 
$\exp(x-s) + \exp(-s) \le 1$,
it follows that the logistics regression loss $\logloss$ 
minimization problem can be formulated as an exponential cone
optimization problem
\begin{subequations}\label{eq: model_1}
\begin{alignat}{4}
 \min_{x \in \mathbb{R}^n}  \logloss = \ \min_{x, s, u, v} \ & \frac{1}{m} \sum_{i=1}^m s_i \notag \\
\ \ \ \text{s.t.} \ \ 
& u_{i} \geq \exp{(-y_i A_i x - s_i)}, \ \  &\;  i \in [m] \notag \\
& v_{i} \geq \exp{(s_i)}, & \;  i \in [m] \notag \\
& u_{i} + v_{i} \leq 1,  & \;  i \in [m] \notag \\
& x \in \mathbb{R}^n, \ s, u, v \in \mathbb{R}^m, \notag
\end{alignat}
\end{subequations}
which is readily solvable by modern conic optimization solvers.

\ignore{

$\logloss$ can be represented by exponential cones, allowing us to solve models for $\eta_{\text{CR}}$ and $\eta_{\text{CC}}$ using commercial solvers.

\begin{subequations}\label{eq: model_1}
\begin{alignat}{4}
& \min  \ \sum_{i\in[m]} t_j + \lambda \sum_{i\in[n]} r_i + \mu \sum_{i\in[n]} z_j & \notag \\
\text{s.t.} & || 2x_j, z_j - r_i ||_2 \leq  z_j + r_i & \;  i \in [n] \notag \\
& k_{1j} \geq exp{(u_j - t_j)}  &\;  j \in [m] \notag \\
& k_{2j} \geq exp{t_j} & \;  j \in [m] \notag \\
& k_{1j} + k_{2j} \leq 1  & \;  j \in [m] \notag \\
& u_j= -y_jA_j^T x & \;  j \in [m] \notag \\
& 0 \leq z_j \leq 1 & \;  i \in [n] \notag
\end{alignat}
\end{subequations}

\begin{subequations}
\begin{alignat}{2}
    \min_{x,z} \;& \logloss + \frac{1}{\gamma} \sumiN x_j^2 + \mu \sumiN z_j \label{eq: obj_reg_mip}\\
    \text{s.t.} \;& x_j(1-z_j) = 0, \quad  j \in [n] \label{eq: constr_reg_mip}\\
    & x \in \mathbb{R}^n, \; z \in \{0, 1\}^n.
\end{alignat}
\end{subequations}
}

Introducing binary indicator variables $z \in \{0, 1\}^n$
to model the $\ell_0$-regularization terms, 
(REG) can be formulated as a mixed-integer conic optimization problem:
\begin{subequations}
\label{eq: persp_reg_mip}
\begin{alignat}{2}
   \eta_{R} =  \min \;& \logloss + \frac{1}{\gamma} \sumiN \frac{x_j^2}{z_j} + \mu \sumiN z_j \\
    \text{s.t.} \;& x_j(1-z_j) = 0, \quad  j \in [n] \label{eq: constr_persp_reg_mip} \\
    & x \in \mathbb{R}^n, \; z \in \{0, 1\}^n. \label{eq: reg_persp_int}
\end{alignat}
\end{subequations}
Here, we adopt the convention 
${x_j^2}/{z_j} = 0 \text{ if } z_j = 0, x_j = 0$ and 
${x_j^2}/{z_j} = \infty  \text{ if } z_j = 0, x_j \neq 0.$
Constraint \eqref{eq: constr_persp_reg_mip} ensures that $x_j = 0$ whenever $z_j = 0$. This constraint can be linearized using the ``big-\emph{M}" technique by replacing it with $-Mz_j \leq x_j \leq Mz_j,$ where $M$ is a large enough positive scalar. However, such big-\emph{M} constraints lead to very weak convex relaxations as we show in the computational experiments in Section~\ref{sec:comp}. 

Instead, we use the conic formulation of the perspective function of $x_j^2$ to model them more effectively. Replacing $x_j^2$ in the objective with its perspective function $x_j^2/z_j$ significantly strengthens the convex relaxation when $0 < z_j < 1$, and introducing $t_j \ge 0$, the perspective can be stated as a rotated second-order cone constraint $x_j^2 \le z_j t_j$ \cite{akturk2009strong}. 
Dropping the complementary constraints \eqref{eq: constr_persp_reg_mip} as well as the integrality constraints on $z$, we arrive at the respective conic (convex) relaxation for (REG): 
\begin{subequations}
\label{eq: persp_reg}
\begin{alignat}{2}
    \eta_{{CR}} = \min \;& \logloss + \frac{1}{\gamma} \sumiN t_j + \mu \sumiN z_j \label{eq: obj_cr_mip}\\
    & x_j^2 \le z_j t_j, \ j \in [n] \label{eq: constr_rotated}  \\
    & x \in \mathbb{R}^n, t \in \mathbb{R_+}^n, z \in [0, 1]^n.
\end{alignat}
\end{subequations}
Note that constraint \eqref{eq: constr_rotated} is valid for $z \in \{0,1\}^n$: as $z_j=0$ implies $x_j=0$ and $z_j=1$ implies simply $x_j^2 \le t_j$, $j \in [n].$

Similarly, one can write the cardinality-constrained version (CARD) as a mixed integer non-linear model with the perspective reformulation:
\begin{subequations}
\label{eq: persp_card_mip}
\begin{alignat}{2}
\eta_{{C}} =   \min \;& \logloss + \frac{1}{\gamma} \sumiN \frac{x_j^2}{z_j} \label{eq: obj_persp_card_mip}\\
    \text{s.t.} \; & \sumiN z_j \leq k \label{eq: constr_persp_card_card_mip} \\
    & x_j(1-z_j) = 0, \quad  j \in [n] \label{eq: constr_persp_card_mip}\\
    & x \in \mathbb{R}^n, z \in \{0, 1\}^n. \label{eq: card_persp_int}
\end{alignat}
\end{subequations}

Dropping \eqref{eq: constr_persp_card_mip} and integrality constraints, and stating the
perspectives as rotated cone constraints, we arrive at the conic relaxation
for (CARD):
\begin{subequations}
\label{eq:  persp_card}
\begin{alignat}{2}
    \eta_{{CC}} = \min \;& \logloss + \frac{1}{\gamma} \sumiN t_j \label{eq: obj_cc_mip}\\
    \text{s.t.} \; & \sumiN z_j \leq k \\
    & x_j^2 \le z_j t_j, \ j \in [n] \\
\  &  x \in \mathbb{R}^n,  t \in \mathbb{R_+}^n, z \in [0, 1]^n.
\end{alignat}
\end{subequations}

\pagebreak
\section{Safe Screening Rules}
\label{sec:screen}
In this section, we first present the safe screening rules for logistic regression with $\ell_0$--$\ell_2$ regularization and then discuss their derivation.

\begin{prop}[Safe Screening Rule for Regularized Logistic Regression (REG)] \label{prop:screen-reg}
Let $x^*$ be an  optimal solution to \perspreg, with objective value $\eta_{{CR}}$, $\alpha_i = y_i/\big (1+ \exp(y_iA_ix^*)\big ),\; i \in [m]$, $ \delta_j = \frac{1}{4}(\alpha' A^{j})^2,\; j \in [n]$, and $\bar{\eta}_{\text{R}}$ be an upper bound on $\eta_R$. Then any optimal solution to \mipreg\ satisfies
$$
z_j =\begin{cases}
	0, & \text{if} \; \eta_{{CR}} + \mu - \gamma \delta_j > \bar{\eta}_{{R}}\\
	1, & \text{if} \; \eta_{{CR}} - \mu + \gamma \delta_j > \bar{\eta}_{{R}}.
    \end{cases}
$$
\end{prop}

\begin{prop}[Safe Screening Rule for Cardinality-constrained Logistic Regression (CARD)] \label{prop:screen-card}
Let $x^*$ be an optimal solution to \perspcard, with objective value $\eta_{{CC}}$,  $\alpha_i = y_i/(1+ exp(y_iA_{i}x^*)), \; i \in [m]$, $ \delta_j = \frac{1}{4}(\alpha' A^{j})^2, \; j \in [n]$, $\delta_{[k]}$ denote the $k\text{-th}$ largest value of $\delta$, and $\bar{\eta}_{{C}}$ be an upper bound on $\eta_C$. Then any optimal solution to \mipcard\ satisfies
$$
z_j =\begin{cases}
	0, & \text{if} \; \delta_j \leq \delta_{[k+1]} \; \text{and} \; \eta_{{CC}} - \gamma(\delta_j - \delta_{[k]}) > \bar{\eta}_{C}\\
	1, & \text{if} \; \delta_j \geq \delta_{[k]} \; \text{and} \; \eta_{{CC}} + \gamma(\delta_j - \delta_{[k+1]}) > \bar{\eta}_{C}.
    \end{cases}
$$

\end{prop}

\subsection{Derivation of Proposition 1}
\label{sec:derive1}

In this section, we present the derivation for the screening rule for (REG) via Fenchel duality. Similar to \citet{atamturk2020safe}, we utilize the dual of the 
perspective terms. In particular, for $p, q \in \mathbb{R}$, consider the convex conjugate, $h^*(p, q)$ of the perspective function $h(x, z) = x^2/z$: 
\begin{equation} 
h^*(p, q) = \max_{x, z} px + qz - \frac{x^2}{z} \cdot \label{eq:convex_conj}
\end{equation}

By Fenchel's inequality, we have $ px + qz - h^*(p, q) \leq \frac{x^2}{z}\cdot$ 
Therefore, for any $p, q \in \mathbb{R}^n$,
we can replace the perspective terms in the objective of \perspreg\ to derive a 
lower bound on $\eta_{CR}$. Then, the Fenchel dual of \perspreg\ is obtained by maximizing the lower bound:
\begin{multline}
   \max_{p,q}  \! \min_{x,z\in[0,1]^n} \logloss + \mu \sumiN z_j \\
  + \frac{1}{\gamma}\Big (p'x+q'z - \sumiN  h^*(p_j, q_j) \Big)  \cdot \label{eq: L_pq}
\end{multline}

Observing that $px + qz - \frac{x^2}{z} $ is concave in $x$ and $z$, allows one to get a closed form solution for \eqref{eq:convex_conj}. Indeed, by simply setting the partial derivatives to zero, we obtain 
\ignore{the following conditions:
 \begin{equation}
 p - \frac{2x}{z} = 0 \label{eq:cond1}
 \end{equation}
 \begin{equation}
     q + \left (\frac{x}{z} \right )^2 = 0 \label{eq:cond2}
 \end{equation}
If these conditions are not satisfied, then $h^*(p,q)$ is unbounded. Hence,}
$$
h^*(p, q) =\begin{cases}
			0, &  q = -p^2/4\\
            		\infty, & \text{otherwise.}
		 \end{cases}
$$
\ignore{
Therefore, it is implied that $q = -\frac{p^2}{q}$, as well as $px + qz - \frac{x^2}{z} = 0$ when this condition holds. The latter is obtained by multiplying \eqref{eq:cond1} by $z$ and summing it with the multiplication of \eqref{eq:cond2} with $x$. 
}

Then, replacing $q_j$ with $-p_j^2/4$ and using the closed form solution for $h^*$, we obtain from \eqref{eq: L_pq} the simplified form of the Fenchel dual:

\begin{align} \label{eq: reg_fd} \! \eta_{{FR}} \! = \!
 \max_{p} \! \min_{x, z\in[0,1]^n} \! \! \logloss \! + \! \! \sumiN \! \Big ( \mu z_j \! + \! \frac{p_j}{\gamma}x_j \! - \! \frac{p_j^2}{4\gamma}z_j \Big).
 \end{align}

Note that \eqref{eq: reg_fd} is concave in $p$. 
Taking the derivative of \eqref{eq: reg_fd} with respect to $p_j$, we obtain the optimal $p_j^* = 2 {x_j}/{z_j}, \; j \in [n]$. 
Plugging $p^*$ into \eqref{eq: reg_fd}, we see that it is equivalent to \eqref{eq: persp_reg}, implying that the dual is tight, i.e.,
$\eta_{CR} = \eta_{FR}$ 

For the inner minimization problem, taking the derivative with respect to $z_j$, we find the optimality conditions 
 		$$ z_j =\begin{cases}
			0, &   \mu - \frac{p^2}{4} > 0 \\
            		1, & \mu - \frac{p^2}{4} < 0. 
		 \end{cases}$$
If $\mu - \frac{p^2}{4} = 0$, then $z_j \in [0,1]$.
On the other hand, taking the derivative with respect to $x_j$ we derive the following optimality condition:
$$\frac{p_j}{\gamma} =  \sum_{i=1}^m \frac{y_i A_{ij}}{1 + \exp(y_iA_ix)} \cdot$$

Let $x^*$ be the optimal solution, and, for $i \in [m]$, define 
$$\alpha_i := y_i/(1+ \exp(y_iA_ix^*)), \ \text{ for } i \in [m]$$
and
$$\delta_j := \frac{1}{4} {(\alpha' A^{j})^2}, \text{ for } j \in [n].$$
Then, $p^* = {\gamma} A^T \alpha$. Furthermore,
$$ \mu - \frac{(p_j^*)^2}{4\gamma} = \mu - \frac{{\gamma} (\alpha' A^{j})^2}{4} = \mu - \gamma\delta_j,$$

Using this closed form solution, we can obtain $p^*$ for \fdr\ from the optimal solution of \perspreg\ via $\alpha$, which in turn can be used to recover $z_j^*, \ j \in [n]$.

\emph{Proof of Proposition 1.} Suppose $\proponepenalty >0$. Then $z_j^* = 0$ in \fdr, and further $\eta_{CR} - (\proponepenalty) < \bar{\eta}_C$. Suppose we add a constraint $z_j = 1$ to \fdr. Let the optimal objective value for this problem be $\etazone$. Since $\eta_{{FR}} + \proponepenalty \leq \etazone$, then if $\eta_{{FR}} + \proponepenalty > \bar{\eta}_R$, there exists no feasible solution for \perspreg\ with $z_j=1$ that has a lower objective than $\bar{\eta}_R$. But, this implies that no optimal solution for \mipreg\ has $z_j=1$, and thus it must be that $z_j=0$.

The same argument is used for the case that $\proponepenalty<0$ and $z_j^* = 1$ in an optimal solution to \fdr. Since $\eta_{{FR}} - (\proponepenalty) = \eta_{{FR}} - \mu + \gamma\delta_j \leq \etazzero$, if $\eta_{{FR}} - \mu + \gamma\delta_j > \bar{\eta}_R$, then the optimal solution for \mipreg\ must have $z_j = 1$.

\subsection{Derivation of Proposition 2}
Using steps similar to in Section~\ref{sec:derive1} we derive the Fenchel dual for \eqref{eq: persp_card}:

\begin{align} \label{eq: card_fd}
\eta_{{FC}} = \text{max}_{p} &\min_{x, z\in[0,1]^n} \logloss + \frac{1}{\gamma} \sumiN \Big (p_jx_j - \frac{p_j^2}{4}z_j \Big ) \nonumber\\
 & \text{s.t.} \sum_{i=1}^n z_j \leq k.
 \end{align}
 
Similarly it can be shown that $p_j^* = 2 {x_j}/{z_j}, \; j \in [n]$, and thus there is no duality gap and $\eta_{{CC}} = \eta_{{FC}}$.
Again, taking the derivative we see that for the minimization problem, the optimal solution for \fdc\ has $z_j = 1$ for the $k$ most negative values of $\mu - \frac{p_j^2}{4}$ which simply translates to the $z_j$ with the $k$ largest values of $\frac{p_j^2}{4}$, with the rest of the indicator variables being equal to zero. In the case that there is no tie between the $k$-th and $(k+1)$-th most largest values, then there is a unique optimal solution for \fdc\ which is integer in $z$, which is therefore the unique optimal solution for \mipcard. Again, we can recover $p^* = {\gamma} A^T \alpha$, and find that $-\frac{(p^*_j)^2}{4\gamma} = -\gamma\delta_j.$

\emph{Proof for Proposition 2.} Suppose $\delta_j \leq \delta_{[k+1]}$. Then $x_j = 0$ in an optimal solution for \fdc. Adding the constraint $z_j=1$, one obtains a solution where the ($k-1$) indicators with the largest values of $\delta$ are set to 1, as well as $z_j$, implying $z_{[k]} = 0$ by the cardinality constraint. But since $\eta_{{FC}} - \gamma\delta_j + \gamma\delta_{[k]} \leq \etaczone$, there exists no optimal solution for \mipcard\ with $z_j = 1$ if $\eta_{{FC}} - \gamma(\delta_j + \delta_{[k]}) > \bar{\eta}_C$.

Using the same argument, if $\delta_j \geq \delta_{[k]}$, then $z_j = 1$ in an optimal solution for \fdc. Adding the constraint $z_j=0$, we obtain a solution with $\delta_{[k+1]} = 1$ as the solution sets the indicator with the next largest $\delta$ to one. Therefore, 
$\eta_{{FC}} + \gamma\delta_j - \gamma\delta_{[k+1]} \leq \etaczzero$, and thus there exists no solution for \mipcard\ with $z_j= 1$ if $\eta_{{FC}} + \gamma(\delta_j - \delta_{[k]}) > \bar{\eta}_C$.
\section{Computational Results}
\label{sec:comp}
In this section, we present the computational experiments performed to test the effectiveness of the safe screening rules described in Section~\ref{sec:screen} for the $\ell_0-\ell_2$ regularized and cardinality-constrained logistic regression problem. 
We test the proposed screening methods on synthetic datasets as well as on real datasets. 

\subsection{Experimental Setup}

The real data instances of varying sizes are obtained from the UCI Machine Learning Repository \cite{dua2017uci} as well as genomics data from the Gene Expression Omnibus Database \cite{edgar2002gene}.

Synthetic datasets are generated using the methodology described in \citet{dedieu2021learning}. Given a number of features $n$ and a number of observations $m$, we generate a data matrix $A \sim \mathcal{N}_n(\boldsymbol{0}, \Sigma)$, and a sparse binary vector $\tilde{x}$, representing the ``true" features, which has $k$ equi-spaced entries equal to one and the remaining entries equal to zero. For each observation $i \in [m]$, we generate a binary label $y_i$, where $Pr(y_i = 1 | A_i) = (1+\exp(-sA_i\tilde{x}))^{-1}$. The covariance matrix $\Sigma$ controls the correlations between features, and $s$ can be viewed as the signal-to-noise ratio. For each experimental setting, we generate ten random instances and report the average of the results for these ten instances for experiments with synthetic data.

We compare the performance of solving (REG) and (CARD)  using \citet{mosek} mixed-integer conic branch-and-bound algorithm with and without screening. For consistency of the runs, we fix the solver options as follows: the branching strategy is set to pseudocost method, node selection is set to best bound method, and presolve and heuristics that add random factors to the experiments are turned off. Upper bounds used for the screening rules are obtained by simply rounding the conic relaxation solution to a nearest feasible integer solution.


\subsection{Results on Synthetic Data}
We first present the experimental results with screening procedure applied to the synthetic datasets. We test the regularized logistic regression (REG) with $n=500, s=1000, k=50$ as a function of the number of observations, $m \in \{200, 500, 1000\}$, the strength of the $\ell_2$ regularization,  $\gamma \in \{1, 1.5, 1.8\}$, and the $\ell_0$ regularization, $\mu \in \{5e^{-4}, 1e^{-3}\}$. For the cardinality-constrained model (CARD), we use the same setting and vary $\gamma$ in the same way while changing the ratio $k/n \in \{0.25, 0.05, 0.017\}$ by fixing $k=50$ and varying $n$. In both experiments, $\Sigma = I$, which corresponds to generating features that are independent of one another.

\begin{figure}
\vskip 0.2in
\begin{center}
\centerline{\includegraphics[width=\columnwidth]{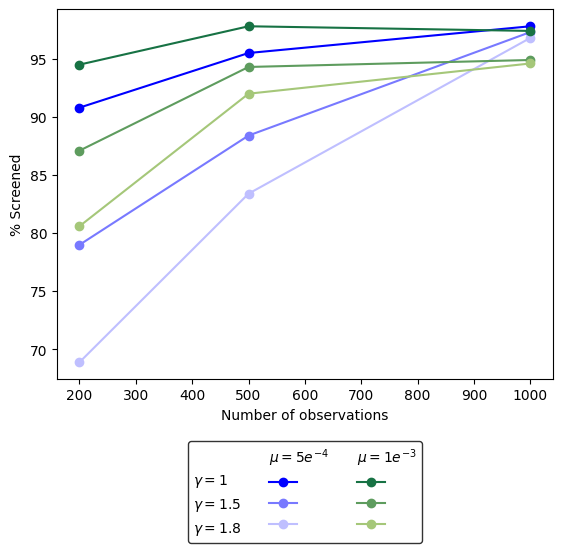}}
\caption{Percentage of features screened as a function of the number of observations in the dataset and regularization strength for (REG).}
\label{fig: reg-screen}
\end{center}
\vskip -0.2in
\end{figure}
\begin{figure}[h]
\vskip 0.2in
\begin{center}
\centerline{\includegraphics[width=\columnwidth]{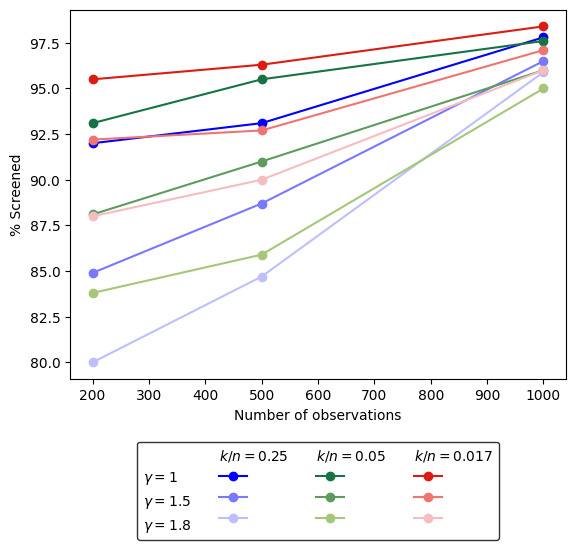}}
\caption{Percentage of features screened as a function of the number of observations in the dataset and regularization strength for (CARD).}
\label{fig: card-screen}
\end{center}
\vskip -0.2in
\end{figure}

Figures \ref{fig: reg-screen} and \ref{fig: card-screen} show the percentage of features eliminated from the regression by the screening procedure for different regularization strengths for (REG) and (CARD), respectively. As the number of observations increases, the number of screened features increases as well. We observe the same trend as the strength of the regularization increases, i.e., higher values of $\mu$ and lower values of $\gamma$ lead to better screening. The reason for improved screening with larger number of observations and stronger regularization can be explained by the smaller integrity gap of the conic relaxations, as shown in Tables \ref{tab: reg-gaps} and \ref{tab: card-gaps}. Integrality gap of a relaxation 
is the relative gap between the optimal objective value of the mixed-integer problem and the relaxation. Smaller integrality gaps lead to the satisfaction of a higher number of screening rules in Propositions~\ref{prop:screen-reg} and \ref{prop:screen-card}.

\begin{table}[h]
\caption{Integrality gap of big-\emph{M} and conic formulations for (REG).}
\vskip 0.15in
\label{tab: reg-gaps}
\centering
\begin{small}
\begin{tabularx}{0.5\textwidth}{XXX|XXXXXc} \hline \hline
                   & \multicolumn{1}{l}{} &  & \multicolumn{3}{c|}{Big-\emph{M} relaxation}      & \multicolumn{3}{c}{Conic relaxation} \\ \hline
$\mu$ &
  \multicolumn{1}{l}{$m$} &
  $\gamma$ &
  \multicolumn{1}{c}{1} &
  \multicolumn{1}{c}{1.5} &
  \multicolumn{1}{c|}{1.8} &
  \multicolumn{1}{c}{1} &
  \multicolumn{1}{c}{1.5} &
  \multicolumn{1}{c}{1.8} \\ \hline
$5e^{-4}$ & 200         &  & 12.91 & 15.06 & \multicolumn{1}{r|}{16.05} & 0.01          & 0.02          & 0.04       \\
                   & 500         &  & 8.43  & 10.18 & \multicolumn{1}{r|}{10.97} & $7e^{-3}$       & 0.02          & 0.03       \\
                   & 1000        &  & 6.00  & 7.15  & \multicolumn{1}{r|}{7.69}  & $7e^{-3}$       & $9e^{-3}$       & 0.01       \\ \hline
$1e^{-3}$ & 200         &  & 15.81 & 18.94 & \multicolumn{1}{r|}{20.34} & 0.02          & 0.04          & 0.06       \\
                   & 500         &  & 9.88  & 12.38 & \multicolumn{1}{r|}{13.51} & 0.01          & 0.03          & 0.04       \\
\textbf{}          & 1000        &  & 7.39  &   9.23  & \multicolumn{1}{r|}{10.01} & 0.01          & 0.03          & 0.03       \\ \hline
\multicolumn{3}{c|}{Average} &
  10.07 &
  12.16 &
  \multicolumn{1}{r|}{13.10} &
  0.01 &
  0.03 &
  0.04 \\ \hline \hline
\end{tabularx}
\end{small}
\end{table}

\begin{table}
\caption{Integrality gap of big-\emph{M} and conic formulations for (CARD).}
\vskip 0.15in
\label{tab: card-gaps}
\centering
\begin{small}
\begin{tabularx}{0.5\textwidth}{XXX|XXXXXX} \hline \hline
\multicolumn{1}{l}{} &
  \multicolumn{1}{l}{} &
   &
  \multicolumn{3}{c|}{\text{Big-\emph{M} relaxation}} &
  \multicolumn{3}{c}{\text{Conic relaxation}} \\ \hline
\multicolumn{1}{l}{$k/n$} &
  \multicolumn{1}{l}{$m$} &
  $\gamma$ &
  \multicolumn{1}{c}{1} &
  \multicolumn{1}{c}{1.5} &
  \multicolumn{1}{c|}{1.8} &
  \multicolumn{1}{c}{1} &
  \multicolumn{1}{c}{1.5} &
  \multicolumn{1}{c}{1.8} \\ \hline
0.250      & 200         &  & 10.23 & 13.32 & \multicolumn{1}{r|}{14.86} & 0.02 & 0.05 & 0.07 \\
\textbf{}            & 500         &  & 5.27  & 7.12  & \multicolumn{1}{r|}{8.08}  & 0.01 & 0.03 & 0.04 \\
\textbf{}            & 1000        &  & 2.87  & 3.91  & \multicolumn{1}{r|}{4.46}  & $4e^{-3}$ & 0.01 & 0.02 \\ \hline
0.050      & 200         &  & 19.49 & 24.33 & \multicolumn{1}{r|}{26.61} & 0.03 & 0.07 & 0.10 \\
\textbf{}            & 500         &  & 10.48 & 13.82 & \multicolumn{1}{r|}{15.51} & 0.01 & 0.03 & 0.05 \\
                     & 1000        &  & 6.14  & 8.25  & \multicolumn{1}{r|}{9.33}  & $6e^{-3}$ & 0.02 & 0.02 \\ \hline
0.017      & 200         &  & 41.74 & 47.70 & \multicolumn{1}{r|}{50.19} & 0.05 & 0.10 & 0.11 \\
\multicolumn{1}{l}{} & 500         &  & 26.58 & 32.73 & \multicolumn{1}{r|}{-}     & 0.02 & 0.06 & -     \\
\multicolumn{1}{l}{} & 1000        &  & 16.67 & 21.47 & \multicolumn{1}{r|}{23.78} & 0.01 & 0.03 & 0.04 \\ \hline 
\multicolumn{3}{c|}{Average} &
   15.50 &
   19.18 &
  \multicolumn{1}{r|}{19.10} &
   0.02&
   0.04&
   0.06\\ \hline \hline
\end{tabularx}
\end{small}
\end{table}

\pagebreak

In Tables \ref{tab: reg-gaps} and \ref{tab: card-gaps}, we also compare the strength of the conic formulation with the big-\emph{M} formulation.
Observe that the integrality gaps produced by the conic relaxation are very small, on average 0.03\% for the regularized model and 0.04\% for the cardinality-constrained model. On the other hand, the big-\emph{M} formulation has a much weaker gap, 12\% and 18\% for the regularized and constrained models, respectively. The tighter gaps with the conic formulation significantly help speed up the solution time of the branch-and-bound algorithm, as well as lead to the elimination of more variables with the screening rules, further speeding up the optimization. 

\begin{table*}[t!]
\caption{Solving times for the regularized logistic regression (REG) with and without screening rules.}
\vskip 0.15in
\label{tab: reg-times}
\centering
\begin{small}
\begin{tabular}{lrl|rrrrrr|rrr} \hline \hline
 &
  \multicolumn{1}{l}{} &
   &
  \multicolumn{6}{c|}{Time (sec.)} &
  \multicolumn{3}{c}{\multirow{2}{*}{Speed-up}} \\
                   & \multicolumn{1}{l}{} &  & \multicolumn{3}{c|}{BnB}                  & \multicolumn{3}{c|}{BnB + Screening} & \multicolumn{3}{c}{}     \\ \hline
$\mu$ &
  \multicolumn{1}{l}{$m$} &
  $\gamma$ &
  \multicolumn{1}{c}{1} &
  \multicolumn{1}{c}{1.5} &
  \multicolumn{1}{c|}{1.8} &
  \multicolumn{1}{c}{1} &
  \multicolumn{1}{c}{1.5} &
  \multicolumn{1}{c|}{1.8} &
  \multicolumn{1}{c}{1} &
  \multicolumn{1}{c}{1.5} &
  \multicolumn{1}{c}{1.8} \\ \hline
$5e^{-4}$ &
  200 &
   &
  16 &
  136 &
  \multicolumn{1}{r|}{264} &
  5 &
  58 &
  127 &
  2.9 &
  2.4 &
  2.1 \\
 &
  500 &
   &
  25 &
  69 &
  \multicolumn{1}{r|}{174} &
  6 &
  19 &
  57 &
  4.3 &
  3.7 &
  3.2 \\
 &
  1,000 &
   &
  30 &
  35 &
  \multicolumn{1}{r|}{49} &
  5 &
  6 &
  9 &
  5.3 &
  5.7 &
  5.5 \\ \hline
$1e^{-3}$ &
  200 &
   &
  10 &
  31 &
  \multicolumn{1}{r|}{69} &
  3 &
  10 &
  25 &
  3.4 &
  3.0 &
  2.9 \\
 &
  500 &
   &
  9 &
  29 &
  \multicolumn{1}{r|}{38} &
  2 &
  6 &
  8 &
  4.2 &
  4.7 &
  4.8 \\
 &
  1,000 &
   &
  39 &
  66 &
  \multicolumn{1}{r|}{71} &
  6 &
  9 &
  10 &
  6.3 &
  7.1 &
  6.7 \\ \hline
\multicolumn{3}{c|}{Average} &
  21 &
  61 &
  \multicolumn{1}{r|}{111} &
  5 &
  18 &
  39 &
  4.4 &
  4.4 &
  4.2 \\
  \hline \hline
\end{tabular}
\end{small}
\end{table*}

\begin{table*}[t!]
\caption{Solution times for the cardinality-constrained logistic regression (CARD) with and without screening rules.}
\vskip 0.15in
\label{tab: card-times}
\centering
\begin{small}
\begin{tabular}{lrl|rrrrrr|rrr} \hline \hline
\multicolumn{1}{l}{} &
  \multicolumn{1}{l}{} &
   &
  \multicolumn{6}{c|}{Time (sec.)} &
  \multicolumn{3}{c}{\multirow{2}{*}{Speed-up}} \\
\multicolumn{1}{l}{} &
  \multicolumn{1}{l}{} &
   &
  \multicolumn{3}{c|}{BnB} &
  \multicolumn{3}{c|}{BnB + Screening} &
  \multicolumn{3}{c}{} \\ \hline
\multicolumn{1}{l}{$k/n$} &
  \multicolumn{1}{l}{$m$} &
  $\gamma$ &
  \multicolumn{1}{c}{1} &
  \multicolumn{1}{c}{1.5} &
  \multicolumn{1}{c|}{1.8} &
  \multicolumn{1}{c}{1} &
  \multicolumn{1}{c}{1.5} &
  \multicolumn{1}{c|}{1.8} &
  \multicolumn{1}{c}{1} &
  \multicolumn{1}{c}{1.5} &
  \multicolumn{1}{c}{1.8} \\ \hline
0.250&
  200 &
   &
  16 &
  40 &
  \multicolumn{1}{r|}{69} &
  4 &
  11 &
  20 &
  4.2 &
  3.6 &
  3.5 \\
\textbf{} &
  500 &
   &
  41 &
  110 &
  \multicolumn{1}{r|}{256} &
  7 &
  23 &
  68 &
  5.8 &
  4.9 &
  4.2 \\
\textbf{} &
  1,000 &
   &
  30 &
  47 &
  \multicolumn{1}{r|}{52} &
  4 &
  7 &
  7 &
  6.7 &
  7.0 &
  7.1 \\ \hline
0.050&
  200 &
   &
  73 &
  200 &
  \multicolumn{1}{r|}{410} &
  12 &
  38 &
  92 &
  6.2 &
  5.5 &
  4.6 \\
\textbf{} &
  500 &
   &
  102 &
  407 &
  \multicolumn{1}{r|}{1,056} &
  13 &
  61 &
  234 &
  8.1 &
  6.8 &
  5.3 \\
 &
  1,000 &
   &
  159 &
  242 &
  \multicolumn{1}{r|}{287} &
  14 &
  23 &
  28 &
  10.8 &
  10.3 &
  10.0 \\ \hline
0.017&
  200 &
   &
  912 &
  2,267 &
  \multicolumn{1}{r|}{1,457} &
  92 &
  1,313 &
  1,703 &
  10.2 &
  8.0 &
  6.4 \\
\multicolumn{1}{l}{} &
  500 &
   &
  1,267 &
  3,548 &
  \multicolumn{1}{r|}{-} &
  167 &
  1,144 &
  1,971 &
  12.6 &
  9.3 &
  - \\
\multicolumn{1}{l}{} &
  1,000 &
   &
  1,166 &
  1,806 &
  \multicolumn{1}{r|}{2,327} &
  57 &
  153 &
  368 &
  19.9 &
  15.3 &
  14.4 \\
  \hline 
  \multicolumn{3}{c|}{Average} &
   418 &
   963 &
  \multicolumn{1}{r|}{740} &
   41 &
   308 &
   499 &
   9.4 &
   7.9 &
   6.9 \\
  \hline
  \hline
\end{tabular}
\end{small}
\end{table*}

In order to see the impact of screening procedure on the overall solution times, we solve the logistic regression problem using the branch-and-bound algorithm with and without screening, and compare the solution times and speed-up due to screening variables. 
The branch-and-bound algorithm for solving the big-\emph{M} formulation exceeds our time limit of 12 hours for the larger instances; therefore, we report results for the perspective formulation only.
These results are shown in Tables \ref{tab: reg-times} and \ref{tab: card-times}. The computation time for the screening procedure is included when reporting the solution times for branch-and-bound with screening. The reported times are rounded to the nearest second. On average, we observe a 4.3$\times$ and 8.1$\times$ speed-up in computations due to the proposed screening procedure for (REG) and (CARD), respectively. The improvement in solution times increases with the number of observations. We continue to see a trend of increased speed-up as the strength of regularization penalty increases, since more features are eliminated a priori.

\subsection{Results on Real Data}

In order to test the effectiveness of the proposed screening procedures on real data, 
we solve problems from the UCI Machine Learning Repository \cite{dua2017uci} (\texttt{arcene} and \texttt{newsgroups}) and genomic data from the Gene Expression Omnibus Database \cite{edgar2002gene} (\texttt{genomic}). In particular, for this experiment, we focus on these larger instances of the repository with a high ratio of features to observations for which regularization is more important to avoid overfitting. We solve these instances using the regularized logistics regression model (REG), varying the strength of the regularization. As before, the time limit is set to 12 hours for each run. 

The results are summarized in Table \ref{tab: real}. For each instance, at least 92\% of the features are screened, and particularly for the genomic dataset, 99.9\% of the features are screened for each parameter setting.
Over all instances, on average, 98\% of the features are eliminated by the screening procedure before the branch-and-bound algorithm. 
Seven out of the 18 runs did not complete in 12 hours without screening. On the other hand,  with screening, all but one run is completed within the time limit and always much faster.
For the instances where branch-and-bound with and without screening both terminate within the time limit, screening leads to on average 13.8$\times$ speed-up, with larger speed-up (up to 25.6$\times$) for the more difficult instances. These experimental results clearly indicate that the proposed screening rules are very effective in pruning a large number of features and result in substantial savings in computational effort for the real datasets as well.

\begin{table*}[t!]
\caption{Results for screening on real datasets using regularized logistic regression (REG).}
\vskip 0.15in
\label{tab: real}
\centering
\begin{small}
\begin{tabular}{l|rr|l|rr|r} \hline \hline
& &  & & 
  \multicolumn{2}{c|}{Time (sec.)} &
  \multicolumn{1}{c}{\multirow{2}{*}{Speed-up}} \\
 &
  $\mu$ &
  $\gamma$ &
  \% Screened &
  BnB &
  BnB + Screening &
  \\ \hline
\multirow{6}{*}{\begin{tabular}[c]{@{}l@{}}\texttt{genomic}\\ $n=22,883$\\ $m= 107$\end{tabular}} &
  \multirow{3}{*}{$5e^{-4}$} &0.5 &
  99.9 &
  104 &
  19 &
  5.5 \\
 &
   &
  1 &
  99.9 &
  182 &
  17 &
  11.0 \\
 &
   &
  1.5 &
  99.9 &
  184 &
  33 &
  5.5 \\ \cline{2-7} 
 &
  \multirow{3}{*}{$1e^{-3}$} &
  0.5 &
  99.9 &
  152 &
  14 &
  11.0 \\
 &
   &
  1 &
  99.9 &
  445 &
  32 &
  13.8 \\
 &
   &
  1.5 &
  99.9 &
  384 &
  54 &
  7.1  \\ \hline \hline
\multirow{6}{*}{\begin{tabular}[c]{@{}l@{}}\texttt{arcene}\\ $n=10,000$\\ $m= 100$\end{tabular}} & \multirow{3}{*}{$5e^{-4}$} & 0.5 & 97 & 25,963 & 1,013 & 25.6 \\
 &
   &
  1 &
  97 &
  6,999 &
  336 &
  20.8 \\
 &
   &
  1.5 &
  92 &
  - &
  10,925 &
  - \\ \cline{2-7} 
 &
  \multirow{3}{*}{$1e^{-3}$} &
  0.5 &
  99 &
  477 &
  32 &
  14.8 \\
 &
   &
  1 &
  96 &
  10,044 &
  467 &
  21.5 \\
 &
   &
  1.5 &
  95 &
  22,466 &
  1,425 &
  15.7    \\ \hline \hline
\multirow{6}{*}{\begin{tabular}[c]{@{}l@{}}\texttt{newsgroups}\\ $n=28,467$\\ $m=1,977$\end{tabular}} &
  \multirow{3}{*}{$5e^{-4}$} &
  0.5 &
  99.9 &
  - &
  1,135 &
  - \\
 &
   &
  0.7 &
  99.9 &
  - &
  8,701 &
  - \\
 &
   &
  1 &
  99 &
  - &
  - &
  - \\ \cline{2-7} 
 &
  \multirow{3}{*}{$1e^{-3}$} &
  0.5 &
  99.9 &
  - &
  401 &
  - \\
 &
   &
  0.7 &
  99.9 &
  - &
  522 &
  - \\
 &
   &
  1 &
  99.7 &
  - &
  7,439 &
  -     \\ \hline \hline
\end{tabular}
\end{small}
\end{table*}

\ignore{
In the \texttt{arcene} and \texttt{newsgroups} datasets, there are instances which cannot be solved (either not completing within the 12 hour time limit or due to numerical instability) without screening, but they are solved within 7--20 minutes when screening is incorporated.
}

\ignore{
\begin{table*}[t]
\caption{Results for the Arcene dataset, which has $N =10,000$ features and $M =100$ observations.}
\vskip 0.15in
\label{tab: arcene}
\centering
\begin{tabular}{rr|llll}
\multicolumn{1}{c}{\textbf{$\mu$}} & \multicolumn{1}{c|}{\textbf{$\gamma$}} & \multicolumn{1}{c}{\textbf{\% Screened}} & \multicolumn{1}{c}{\textbf{BnB}} & \multicolumn{1}{c}{\textbf{Bnb + Screen}} & \multicolumn{1}{c}{\textbf{Speed-up}} \\ \hline
0.001                              & 0.5                                    & 99                                      & 477                                & 32                                          & 14.84                                \\
0.001                              & 1                                      & 96                                      & 10044                              & 467                                         & 21.49                                \\
0.001                              & 1.5                                    & 95                                      & 22466                              & 1425                                        & 15.77                                \\ \hline
0.0005                             & 0.5                                    & 97                                      & 25963                              & 1013                                        & 25.63                                \\
0.0005                             & 1                                      & 97                                      & 6999                               & 336                                         & 20.82                                \\
0.0005                             & 1.5                                    & 92                                      & \multicolumn{1}{c}{-}                 & 10925                                       & \multicolumn{1}{c}{-}               
\end{tabular}
\end{table*}
}
\ignore{
\begin{table*}[ht]
\caption{Results for the Genomic dataset, which has $N = 22,883$ features and $M = 107$ observations.}
\vskip 0.15in
\label{tab: genomic}
\centering
\begin{tabular}{rr|llll}
\multicolumn{1}{c}{\textbf{$\mu$}} & \multicolumn{1}{c|}{\textbf{$\gamma$}} & \multicolumn{1}{c}{\textbf{\% Screened}} & \multicolumn{1}{c}{\textbf{BnB}} & \multicolumn{1}{c}{\textbf{Bnb + Screen}} & \multicolumn{1}{c}{\textbf{Speed-up}} \\ \hline
0.001                              & 0.5                                    & 99.9                                     & 152                                & 14                                             & 11.04                                \\
0.001                              & 1                                      & 99.9                                     & 445                                & 32                                            & 13.83                                \\
0.001                              & 1.5                                    & 99.9                                     & 384                                & 54                                            & 7.12                                 \\ \hline
0.0005                             & 0.5                                    & 99.9                                     & 104                                & 19                                            & 5.46                                 \\
0.0005                             & 1                                      & 99.9                                     & 182                                & 17                                            & 11.01                                \\
0.0005                             & 1.5                                    & 99.9                                     & 184                                & 33                                            & 5.53                                
\end{tabular}
\end{table*}
}
\ignore{
\begin{table*}[ht]
\caption{Results for the newsgroups dataset, which has $N = 28,467$ features and $M = 1977$ observations.}
\vskip 0.15in
\label{tab: newsgroups}
\centering
\begin{tabular}{rr|llll}
\multicolumn{1}{c}{\textbf{$\mu$}} & \multicolumn{1}{c|}{\textbf{$\gamma$}} & \multicolumn{1}{c}{\textbf{\% Screened}} & \multicolumn{1}{c}{\textbf{BnB}} & \multicolumn{1}{c}{\textbf{Bnb + Screen}} & \multicolumn{1}{c}{\textbf{Speed-up}} \\ \hline
0.001                              & 0.5                                    & 99.9                                     & -                                & 401                                             & -                                \\
0.001                              & 0.7                                      & 99.9                                     & -                                & 522                                            & -                                \\
0.001                              & 1                                    & 99.7                                     & -                                & 7439                                            & -                                 \\ \hline
0.0005                             & 0.5                                    & 99.9                                     & -                                & 1135                                            & -                                 \\
0.0005                             & 0.7                                      & 99.7                                     & -                                & 8701                                            & -                                \\
0.0005                             & 1                                    & 99                                     & -                                & -                                            & -                                
\end{tabular}
\end{table*}
}

\section{Conclusion}
\label{sec:conc}
In this work, we present safe screening rules for $\ell_0-\ell_2$ regularized and cardinality-constrained logistic regression. Our numerical experiments show that a large percentage of features can be eliminated efficiently and safely via this preprocessing step before employing branch-and-bound algorithms, particularly when regularization is strong, leading to significant computational speed-up. The strength of the conic relaxations contribute significantly to the effectiveness of the screening rules in pruning a large number of features. We show the conic formulation provides much smaller integrality gaps compared to the big-\emph{M} formulation, making it more suitable for solving $\ell_0$--$\ell_2$-regularized logistic regression with a branch-and-bound algorithm and also for the derived screening rules.


\bibliography{main}
\bibliographystyle{icml2022}

\end{document}